\documentclass[times,twocolumn,final,authoryear]{elsarticle}

\usepackage{prletters}
\usepackage{framed,multirow}

\usepackage{amssymb}
\usepackage{latexsym}

\usepackage{url}
\usepackage{xcolor}

\usepackage{blindtext, graphicx}

\usepackage[utf8]{inputenc} 
\usepackage[T1]{fontenc}    
\usepackage{hyperref}       
\usepackage{url}            
\usepackage{booktabs}       
\usepackage{amsfonts}       
\usepackage{nicefrac}       
\usepackage{microtype}      
\usepackage{blindtext, graphicx}
\usepackage[utf8]{inputenc}
\usepackage[T1]{fontenc}
\usepackage{array,booktabs}
\usepackage{amsmath,cite}

\definecolor{newcolor}{rgb}{.8,.349,.1}

\begin{document}

\ifpreprint
  \vspace*{-1pc}
\else
\fi

\ifpreprint
  \setcounter{page}{1}
\else
  \setcounter{page}{1}
\fi

\begin{frontmatter}

\title{Order embeddings and character-level convolutions for multimodal alignment}

\author[1]{Jônatas \snm{Wehrmann}}
\author[1]{Anderson \snm{Mattjie}}
\author[1]{Rodrigo C. \snm{Barros}\corref{cor1}}
\cortext[cor1]{Corresponding author: 
  Tel.: +55-51-3353-8641;}
\ead{rodrigo.barros@pucrs.br}

\address[1]{Pontifícia Universidade Católica do Rio Grande do Sul, Av. Ipiranga, 6681, Porto Alegre, RS, 90619-900, Brazil}

\received{1 May 2013}
\finalform{10 May 2013}
\accepted{13 May 2013}
\availableonline{15 May 2013}
\communicated{S. Sarkar}

\begin{abstract}
With the novel and fast advances in the area of deep neural networks, several challenging image-based tasks have been recently approached by researchers in pattern recognition and computer vision. In this paper, we address one of these tasks, which is to match image content with natural language descriptions, sometimes referred as multimodal content retrieval. Such a task is particularly challenging considering that we must find a semantic correspondence between captions and the respective image, a challenge for both computer vision and natural language processing areas. For such, we propose a novel multimodal approach based solely on convolutional neural networks for aligning images with their captions by directly convolving raw characters. Our proposed character-based textual embeddings allow the replacement of both word-embeddings and recurrent neural networks for text understanding, saving processing time and requiring fewer learnable parameters. Our method is based on the idea of projecting both visual and textual information into a common embedding space. For training such embeddings we optimize a contrastive loss function that is computed to minimize order-violations between images and their respective descriptions. We achieve state-of-the-art performance in the largest and most well-known image-text alignment dataset, namely Microsoft~COCO, with a method that is conceptually much simpler and that possesses considerably fewer parameters than current approaches.

\end{abstract}

\end{frontmatter}



\section{Introduction}
To learn proper feature representation of input data is an essential part of any machine learning problem, as it directly impacts in the precision of the generated data-based models. Thanks to the fast pace in which computer hardware has been evolving in the last decade, along with the rapid development of computer vision and natural language processing technologies, great advances have been made in tasks that require a huge amount of computational power, in particular those that benefit from optimization-based approaches such as deep neural networks. One of these tasks, image-text alignment, has become an important problem in the latest years as it has many applications such as image and video retrieval, captioning, navigation for the blind, and many others. By successfully mapping image-to-text and text-to-image, we can significantly impact the general and broad task of information retrieval. 

For this multimodal content retrieval task, state-of-the-art results \citep{m_rnn,show_and_tell,noc} rely on either Long Short-Term Memory networks (LSTMs) \citep{Hochreiter1997} or Gated Recurrent Units (GRUs) \citep{Chung2015} with word-embeddings \citep{Mikolov2013}. Although such an approach has shown promising results, it has some drawbacks that are important to consider. First of all, it is costly due to the need of pre-training a word-embedding with a latent space informative enough to capture semantic relationships among words. Second, it takes a considerable amount of storage and memory for dealing with these word embeddings, depending on the size of the dictionary, in which often larger is better in terms of predictive accuracy. 

By taking into consideration the aforementioned drawbacks, we propose a model that, instead of relying in similar recurrent LSTM/GRU-based architectures that depend on pre-trained word-embeddings, learns from scratch, character per character, how to retrieve captions from images and images from captions by making use of convolutional layers alone. Additionally, our model makes no assumptions regarding specific templates, guidelines, or classifications since it learns everything from scratch using the training data. Since image captioning can be seen as a special case of a single visual-semantic hierarchy over words, sentences, and images, we design a loss function based on the so-called order embeddings \citep{vendrov2015order}. This particular type of embeddings are designed for explicitly modeling the partial order structure of the visual-semantic hierarchy existent in image captions. While typical approaches for image captioning rely on mapping words to vectors in a distance-preserving fashion \citep{Socher2014}, we believe order embeddings to be more suitable since the resulting mapping is not distance-preserving but order-preserving between the hierarchy and a partial order over the embedding space, making it easier to relate the naturally-hierarchical concepts within image captions.

In order to evaluate the performance of our model, we execute a series of experiments where we make particular architectural changes to the convolutional neural network by increasing the number of convolutional layers up to 5 and by changing the number of filters. We compare our proposed approach with the current state-of-the-art, and we show that our method achieves state-of-the-art results while often presenting a much lighter, simpler, and easier-to-train architecture.

This paper is organized as follows. Section~\ref{sec:method} presents in detail our novel approach for multimodal content retrieval. Section~\ref{sec:methodology} describes the methodology that we adopt for performing the experimental analysis, and the results are discussed in Section~\ref{sec:results}. Section~\ref{sec:rw} reviews related work, and we end this paper with our conclusions and future work directions in Section~\ref{sec:conclusions}.

\section{Character-based Language Embeddings}\label{sec:method}

The use of word-embeddings~\citep{Mikolov2013} for text understanding has become a standard approach, being largely used across several tasks such as text classification, machine translation, image and video captioning, and information retrieval. Such an approach consists of representing a given word into a multidimensional latent space, $w \in \mathcal R^d$. These embeddings are often projected into a distance-preserving Euclidean space, in which semantic regularities can be easily identified and even manipulated. An example of manipulation of this distance-preserving word-embedding space is the following vector operation over the corresponding words: 
\begin{equation}
    king\ - man\ + woman \sim queen
\end{equation}

Current state-of-the-art sentence embedding approaches  \citep{SkipThoughtVectors,vendrov2015order,mesnil2013investigation,kiros2014visual,karpathy2015deep} have demonstrated similar results when using RNNs and word-embeddings for encoding entire sentences into a $d$-dimensional embedding space. Nevertheless, it is important to emphasize that even though word-embeddings are employed in the current state-of-the-art approaches, they tend to present three major drawbacks: i)~they require pre-training word-embeddings or RNNs in very large corpuses (with millions or billions of words), which demands both time and computational power; ii)~in order to encode a single word or sentence, it is necessary to have ``at hand" the whole word-dictionary containing all known words, which largely increases the memory requirements to store all data; iii)~for multilingual or informal domains (e.g., twitter and internet searches) the number of words in the dictionary increases according to the number of languages and, in addition, preprocessing is often employed for correcting typos and standardizing the words. 

In this work, we propose a novel architecture for learning textual embeddings based on convolving raw characters. Our approach is designed to be simple, efficient, and fast, though still capable of generating state-of-the-art results. Our hypothesis is that a convolutional layer is capable of learning a proper latent embedding space for encoding text semantics. Hence, we replace the word dictionary by applying $f$ convolutional filters over the input text in a temporal window of size $t$. Moreover, to keep $t$ unchanged through the layer computation, we perform padded convolutions. The resulting feature map  $\in \mathcal R ^ {t \times f}$ can be seen as a text-embedding where each character is projected onto a $f$-dimensional space by considering its relation to the adjacent characters (defined by the filters' length). Such an approach does not require fixed-sized dictionaries nor preprocessing of any kind, and its complexity does not raise with the complexity or availability of text. 

\subsection{Character-based Convolutions}

The core idea behind our approach is to replace the use of both RNNs and word dictionaries by applying convolutions to learn textual embeddings. A given text is represented by $\mathcal T \in \{0, 1\} ^ {n \times a}$, where $n$ is the number of characters in the text and $a$ is the alphabet size. Note that $n$ is variable and changes according to the available text, whereas $a$ must be unchanged given that the alphabet contains all known characters. We convolve characters of the input text $\mathcal T$ by applying $f$ convolutional filters of length $l$, where the $j^{th}$ filter in the $i^{th}$ convolutional layer generates feature map $\mathcal F_{ij}$ whose $x^{th}$ position is given by : 

\begin{equation}
    \mathcal{F}^{x}_{ij} =\phi\left(b_{ij}+\sum_{m=0}^{f_{i-1}}\sum_{p=0}^{l-1}w^{p}_{ijm}\mathcal F^{(x+p)}_{(i-1)m}\right)
    \label{eq:convolution}
\end{equation} where $\phi$ is an activation function, $b_{ij}$ is the bias for the respective convolutional filter, $m$ iterates over the feature maps (channels), $p$ indexes the position of the kernel, $w_{ijm}^p$ is the filter weight and $\mathcal F_{(i-1)m}^{x+p}$ is the value of the previous feature map (or input). Note that $m$ iterates over the alphabet size for the case of the first convolution, and over the $f_i$ feature maps of the previous layer for the subsequent convolutions.

A known restriction of applying a single convolutional layer for embedding texts is the size of the receptive field. A convolutional filter of length $l=7$ is capable of learning information of $7$ neighboring characters. This size is probably enough for learning word-based information. Standard strategies for allowing the global learning of the whole text include: i)~increasing $l$, which leads to an exponential growth of parameters, eventually making the learning unfeasible for large-size texts; ii)~adding more convolutional layers, hence requiring more processing resources and parameters depending on the number of filters of the subsequent convolutions; and iii)~using local or global pooling layers. In order to keep our architecture compact, we explore up to five convolutional layers and a maximum filter size of $7$. 

\begin{figure*}[!htpb]
       \centering
       \includegraphics[scale=0.5]{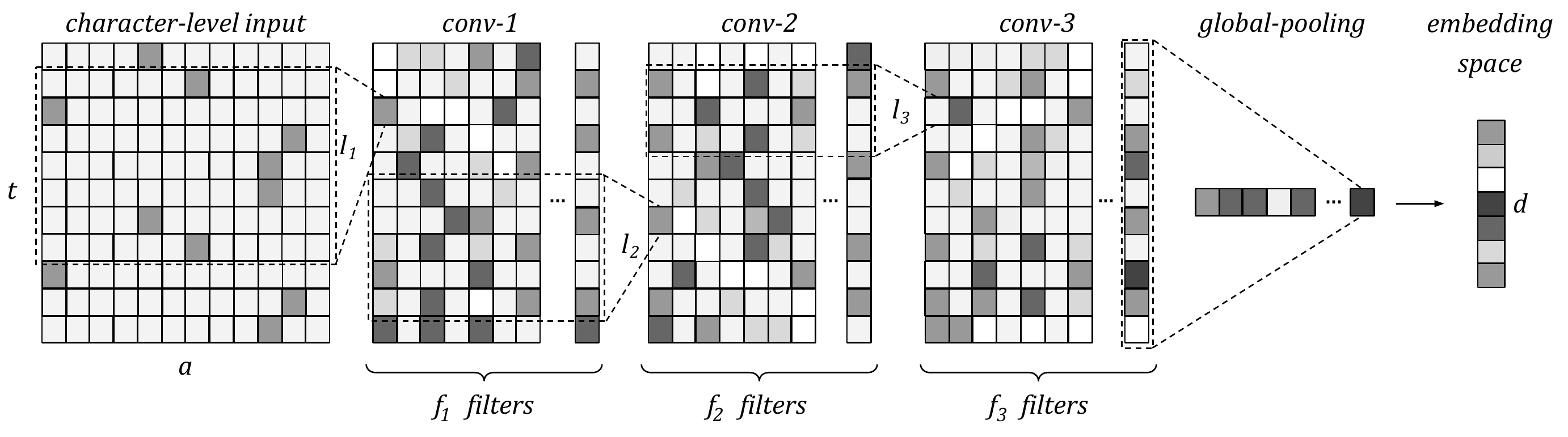}
       \caption{Convolutions-through-time (CTT) module with 3 convolutional layers for processing a text of $t$ characters based on an alphabet of $a$ symbols.}
       \label{fig:ctt}
\end{figure*}

The final text-embedding vectors are generated by applying a max-pooling-over-time layer, which selects the most important features across the temporal dimension of feature map $\mathcal F$. Note that the number of filters in the last convolutional layer defines the length of the embedding vector. Figure~\ref{fig:ctt} presents a schematic of text-embedding via character-based convolutions, which we have named ``convolutions-through-time" (CTT).

\subsection{Architecture}

Our architecture is designed to approximate two encoding functions, $f_t(\mathcal T)$ and  $f_i(\mathcal I)$, whose goal is to project both text $\mathcal T$ and image $\mathcal I$ into the same embedding space. In such a space, correlated image-text pairs should be close to each other, and the distance of non-correlated pairs should necessarily be larger than the correlated ones. For the text encoding function $f_t(\mathcal T)$, we make use of the CTT module described in the previous section.
For the image encoding function $f_i(\mathcal I)$, we extract image features from the second fully-connected layer of a VGG-19 network~\citep{Simonyan2014} pre-trained in the ImageNet dataset~\citep{Russakovsky2015}. For better feature representation, we use the $10$-crop strategy: we scale to $256$-pixels the smallest image size, and sample $224\times224$ crops from the corners, center, and horizontal mirroring. Finally, features from all crops are averaged element-wise. 

Let $\mathcal C (\mathcal I)$ be features extracted from image $\mathcal I$ by the convolutional neural network. Images are projected onto the $\mathcal R^d_+$ embedding-space based on a linear mapping:

\begin{equation}
    f_i(\mathcal I) = |W_i \cdot \mathcal C(\mathcal I)|
\end{equation} where $W_i \in \mathcal R^{d \times 4096}$ is a learned weight matrix and $d$ is the number of dimensions of the embedding space. 

For embedding text, we use the proposed character-based approach $f_t(\cdot)$ with four main variations shown in Table~\ref{tab:architectures}. We use maxout-based convolutions (Eq.~(\ref{eq:maxout})), given that it performs universal approximation of proper activation functions, addressing the rectifier linear unit (ReLU) issue of saturating in the negative region of activations:  

\begin{equation}
    \mathcal F_{m} = max\{\psi(W_1, x), \psi(W_2, x)\}
    \label{eq:maxout}
\end{equation} where $\mathcal F_m$ is a feature map generated by convolving -- $\psi(.)$ -- a given input $x$ with two distinct weight tensors, namely $W_1$ and $W_2$. Each activation value in the output map is selected as the largest of the two values generated by the distinct convolutions.

\begin{table}[!htbp]
\scriptsize
\centering
\caption{Four versions of our approach.}
\begin{tabular*}{\columnwidth}
    {@{\extracolsep{\fill}}clcccc}
\toprule
Version & Architecture  & Output &  \#Filters  ($f$) &  Filter Length ($l$)  & \#Params  \\
\toprule
&       &     &                  &             &                  \\ 
& Input         & $t \times 72$    &                  &              &                   \\ 
A & MaxConv 1     & $t \times 512$   & 512              & 7             & 517,120  \\ 
& Pooling       & $512$            & -                & $t$           & -                     \\
\midrule
&          &                &                &              & 0.52M                    \\ 
\bottomrule
&       &     &                  &             &                  \\ 
& Input         & $t \times 72$    &                  &              &                   \\ 
& MaxConv\_1     & $t \times 256$   & 256              & 7             & 258,560  \\ 
B & MaxConv\_2     & $t \times 512$   & 512              & 5             & 1,311,744 \\ 
& Pooling       & $512$            & -                & $t$           & -                     \\
\midrule
&          &                 &                 &              & 2.10M             \\ 
\bottomrule
&       &     &                  &             &                  \\ 
& Input         & $t \times 72$    &                  &              &                   \\ 
& MaxConv\_1     & $t \times 128$   & 128              & 7             & 129,280  \\ 
C & MaxConv\_2     & $t \times 256$   & 256              & 5             & 328,192 \\ 
& MaxConv\_3     & $t \times 512$   & 512              & 3             & 787,456 \\ 
& Pooling       & $512$            & -                & $t$           & -                     \\
\midrule
&          &                 &                &             & 1.24M               \\
\bottomrule
&       &     &                  &             &                  \\ 
& Input         & $t \times 72$    &                   &              &                   \\ 
& MaxConv\_1     & $t \times 512$   & 512              & 7            &   517,120  \\ 
D & MaxConv\_2     & $t \times 512$   & 512              & 5            & 2,622,464 \\ 
& MaxConv\_3     & $t \times 512$   & 512              & 3            & 1,573,888 \\ 
& Pooling       & $512$             & -                 & $t$           & -                     \\
\midrule
&          &                 &                &            & 4.713M                     \\ 
\bottomrule
\end{tabular*}
\label{tab:architectures}
\end{table}

Since it uses a larger number of parameters, maxout convolutions generate more compact and informative feature maps while saving some memory when compared to ReLU-based networks with the same amount of parameters. In order to keep a reduced amount of parameters, we decreased the number of filters in the shallower convolutional layers, except for Architecture D, which contains the same number of filters in all layers. 

All models are standardized to provide a 512-long vector representation that carries the textual semantic information. Similarly to $f_i(\cdot)$, we linearly project such representation onto $\mathcal R^{d}_+$ by using a learned $W_t \in \mathcal R ^ {d \times 512}$ weight matrix.

Note that the current state-of-the-art approach for image-text alignment~\citep{vendrov2015order} makes use of GRU networks fed with word embeddings, requiring $\approx 8M$ parameters for the word-embeddings and $\approx 4M$ parameters for the GRU itself. Our largest architecture contains $\approx 4.7M$ parameters (almost $3\times$ lighter) and our smallest architecture contains only half a million parameters, $24 \times$ lighter than the state-of-the-art.



\begin{table*}[!htbp]
\scriptsize
\centering
\caption{Bidirectional results in MS COCO test set. Values of the first section are reported in the 1-fold test set. Values of both second (state-of-the-art) and third (our models) sections are averages of the 5-fold test set. Values in bold depict the current state-of-the-art results. Values underlined outperform the best published results.}
\begin{tabular*}{\textwidth}
    {@{\extracolsep{\fill}}lcccccccc}
\toprule
& \multicolumn{4}{c}{Image to text} & \multicolumn{4}{c}{Text to image} \\
\midrule
Method          	     &  R@1 & R@10 & Med $r$ & Mean $r$ & R@1 & R@10 & Med $r$ & Mean $r$  \\
\midrule
MNLM~\citep{kiros2014visual}            	     & 43.4 & 85.8 & 2.0 & *    & 31.0 & 79.9   & 3.0 & *    \\
DVSA~\citep{karpathy2015deep}            	     & 38.4 & 80.5 & \textbf{1.0} & *    & 27.4 & 74.8   & 3.0 & *    \\
FV~\citep{fv2015}            	         & 39.4 & 80.9 & 2.0 & 10.4 & 25.1 & 76.6   & 4.0 & 11.1 \\
m-CNN~\citep{mcnn2015}            	     & 38.3 & 81.0 & 2.0 & *    & 27.4 & 79.5   & 3.0 & *    \\
m-CNN$_{ENS}$~\citep{mcnn2015}            & 42.8 & 84.1 & 2.0 & *    & 32.6 & 82.8   & 3.0 & *    \\
combine-skip-thought~\citep{SkipThoughtVectors}     & 33.8 & 82.1 & 3.0 & *    & 25.9 & 74.6   & 4.0 & *    \\
\midrule
Order-embeddings (symm.)~\citep{vendrov2015order} & 45.4& 88.7 & 2.0 & 5.8  & 36.3 & 85.8   & 2.0 & 9.0  \\
Order-embeddings~\citep{vendrov2015order}         & 46.7 & \textbf{88.9} & 2.0 & 5.7  & \textbf{37.9} & 85.9   & 2.0 & 8.1  \\
\midrule
Arch. A [ours]          & 42.1       & 87.2           &            2.0 &  6.7         &  34.5             & 85.0              &  2.6       &  8.2  \\
Arch. B [ours]          & 46.2       & 88.0           &            2.0 &  5.9         &  36.9             & \underline{86.4}               &  2.0       &  \underline{7.4}  \\
Arch. C [ours]          & 46.2       & 88.6           &            2.0 &  5.8         &  37.0             & \underline{86.6}  &  2.0       &  \underline{7.6}  \\
Arch. D [ours]          & \underline{\textbf{47.2}}       & \textbf{88.9} & 2.0 & \underline{\textbf{5.6}} &  37.5 & \underline{\textbf{87.0}}     &  2.0       &  \underline{\textbf{7.3}}   \\
\bottomrule
\end{tabular*}
\label{tab:results}
\end{table*}

\subsection{Loss function}

Let $f_t(\mathbf{\mathcal T}) = \bf{m}$ be the sentence embedding vector and $f_i(\mathbf{\mathcal I}) = \bf{v}$ be the image embedding. We first scale $\bf{m}$ and $\bf{v}$ to have unit norm, so that the inner product of both results in the cosine distance. Instead of directly optimizing the cosine distance, we decided for optimizing the alignment preserving the order relationships among the visual-semantic hierarchy, given that asymmetric distances are naturally more well-suited for image-sentence alignment. Hence, we apply an order-violation constraint by penalizing an ordered pair $(x, y)$ of points in $\mathcal R^N_+$:

\begin{equation}
    s(x, y) = - \lvert max\{ 0, y-x \}\rvert ^2
\end{equation}

The order violation penalties are used as a similarity distance, and optimized by the following constrastive pairwise ranking loss: 

\begin{multline} \label{eq:loss}
 \mathcal L = \sum_\mathbf{m} \sum_{k}max\{ 0, \alpha - s(\mathbf{m}, \mathbf{v}) + s(\mathbf{m}, \mathbf{v}_k) \} \\ + \sum_\mathbf{v} \sum_k max\{ 0, \alpha - s(\mathbf{v}, \mathbf{m}) + s(\mathbf{v}, \mathbf{m}_k) \}
\end{multline} where $\mathbf{m_k}$ and $\mathbf{v_k}$ are respectively the sentence and image contrastive examples (i.e., uncorrelated). This loss function encourages the similarity $s(x, y)$ for actual image-text pairs to be larger than the contrastive ones by a margin of at least $\alpha$.

\section{Experimental Setup}\label{sec:methodology}


\subsection{Dataset}

For analyzing the performance of our proposed approach, we make use of the Microsoft COCO dataset \citep{coco}. It contains over 100,000 images with at least 5 captions per image, where more than 65\% of the captions are written in English. We have used the same data splits from \citep{karpathy2015deep}: 113,287 images for training, 5,000 images for validation, and 5,000 images for testing. This dataset has been extensively employed in the recent years for image-text retrieval challenges.

\subsection{Hyper-Parameters}

We use the performance on validation data as a proxy for choosing the best hyper-parameters, and we perform a non-exhaustive hyper-parameter grid search. We employ Adam \citep{Kingma2014} for optimization, given its capacity in adjusting per-weight learning rates during training. It performs a learning rate annealing in each iteration based on estimates of first and second-order moments, leading to a deeper exploration of the search space. We use Adam's default initial learning rate as suggested in \citep{Kingma2014}. In addition, we found it was beneficial to reduce the learning rate by $10 \times$ whenever the validation error \textit{plateaus}. Inspired in~\citep{vendrov2015order}, we use a batch size of $100$ (hence, $99$ contrastive examples) and margin $\alpha = 0.05$. Note that neither weight decay nor dropout were used, since we believe the loss function itself is enough to regularize the model by including several contrastive examples instead of using only hard-contrastive ones. 


\subsection{Evaluation Measures}

For evaluating the results, we use the same measures as those in \citep{vendrov2015order}: $R@K$ (reads ``Recall at $K$") is the percentage of queries in which the ground-truth term is one of the first $K$ retrieved results. The higher its value, the better. We also show the results of  \textit{Med}~$r$ and \textit{Mean}~$r$, which represent respectively the median and mean of the ground-truth ranking. Since they are ranking-based measures, the smaller their values the better.

\section{Results}\label{sec:results}

All recent approaches developed for image-text alignment so far make use of word-embeddings, whilst ours is the first to generate and use character-level embeddings for this kind of task. Our results are compared to the state-of-the-art approaches that rely on word-embeddings, namely: MNLM~\citep{kiros2014visual}, DVSA~\citep{karpathy2015deep}, FV~\citep{fv2015}, $m$-CNN~\citep{mcnn2015}, $m$-CNN$_{ENS}$~\citep{mcnn2015}, combine-skip-thought~\citep{SkipThoughtVectors}, order-embeddings and order-embeddings (symm.)~\citep{vendrov2015order}. 

\subsection{Quantitative Analysis}

Table~\ref{tab:results} depicts the results obtained for our 4 architectures along with the published results from the baselines. We first analyze the performance of all methods with respect to the image-to-text task. Note that our approach (Arch.~D) outperforms the previous state-of-the-art by $0.5$ regarding $R@1$, whereas for $R@10$ we obtain the same performance ($88.9$) as the method in~\citep{vendrov2015order}. Considering the ranking measures, the best approach according to the median ranking is DVSA~\citep{karpathy2015deep}, but note that such a result is not directly comparable to ours since they use a single fold as test set whereas we present the average results for all 5 folds like~\citep{vendrov2015order}. With respect to the mean ranking, we once again achieve state-of-the-art results with the lowest ranking of all ($5.6$). If we take into consideration that the architecture in \citep{vendrov2015order} is more complex than ours, besides relying on word-embeddings and hence requiring much more computational resources, we can see that our approach is clearly preferred over the method in~\citep{vendrov2015order}, arguably establishing itself as the novel state-of-the-art for image-to-text retrieval.

Next, we analyze the performance of all methods with respect to the task of text-to-image retrieval. In this task, we only fail at outperforming the state-of-the-art regarding $R@1$, in which our best architecture is outperformed by a margin of 0.4 to~\citep{vendrov2015order}. However, three of our architectures (Arch.~B, C, and D) outperform the state-of-the-art considering $R@10$, with our best architecture reaching 87\% of recall against 85.9\% of the best baseline. The same behavior is verified in the mean ranking, where our best approach (Arch.~D) reaches an average ranking of 7.3 versus 8.1 of the current state-of-the-art. Hence, we come to the same conclusion as before: our approach, besides simpler and lighter, is also often better than the previously-published approaches, hence establishing itself as the novel state-of-the-art for text-to-image retrieval.  

As a secondary note, let us consider Arch.~A, our smallest architecture. Even though it did not achieve the best results regarding the 4 architectures we propose, it still reaches quite competitive results, which is \textit{per se} quite surprising considering that it comprises a single convolutional layer and at least fifteen times fewer parameters than the the architecture in \citep{vendrov2015order}. In addition, this very simple model is actually capable of outperforming the following more complex methods: i)~combine-skip-thoughts, which had to be pre-trained for an entire month in a huge corpora and makes use of three GRUs with 1024 units each for extracting features from texts; ii)~DVSA, which employs R-CNN based features from 20 regions in the images (i.e., features extracted from networks that were trained to perform object detection and localization), a much more complex and computationally demanding approach when compared to simply extracting traditional features from image-classification based networks; iii)~FV, that makes use of Fisher Vectors handcrafted features for text encoding; and iv)~$m$-CNN and its ensemble version, which are CNN-based architectures that learn similarity metrics.



\begin{figure}[!htpb]
       \centering
       \includegraphics[scale=0.54]{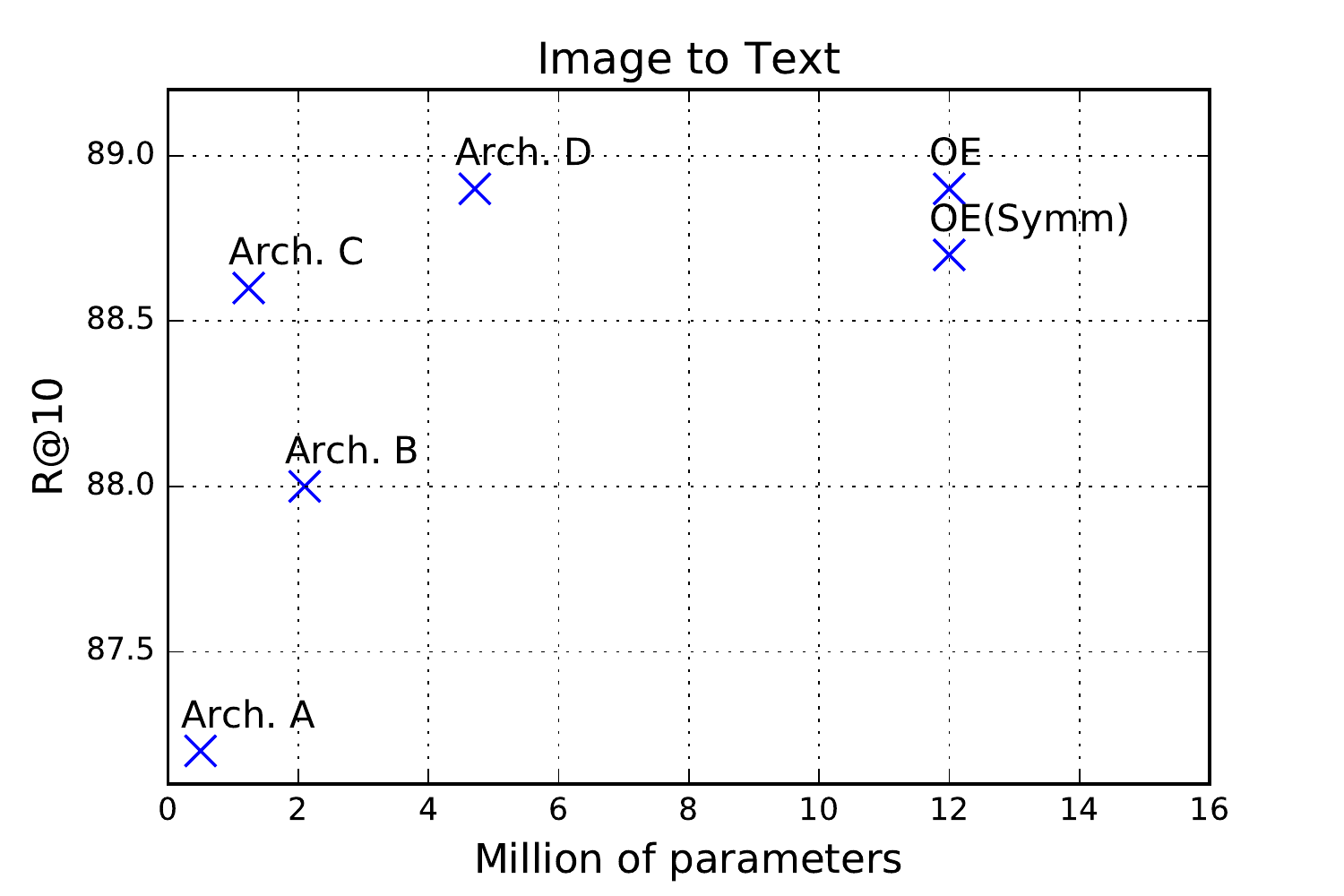} \qquad
       \includegraphics[scale=0.54]{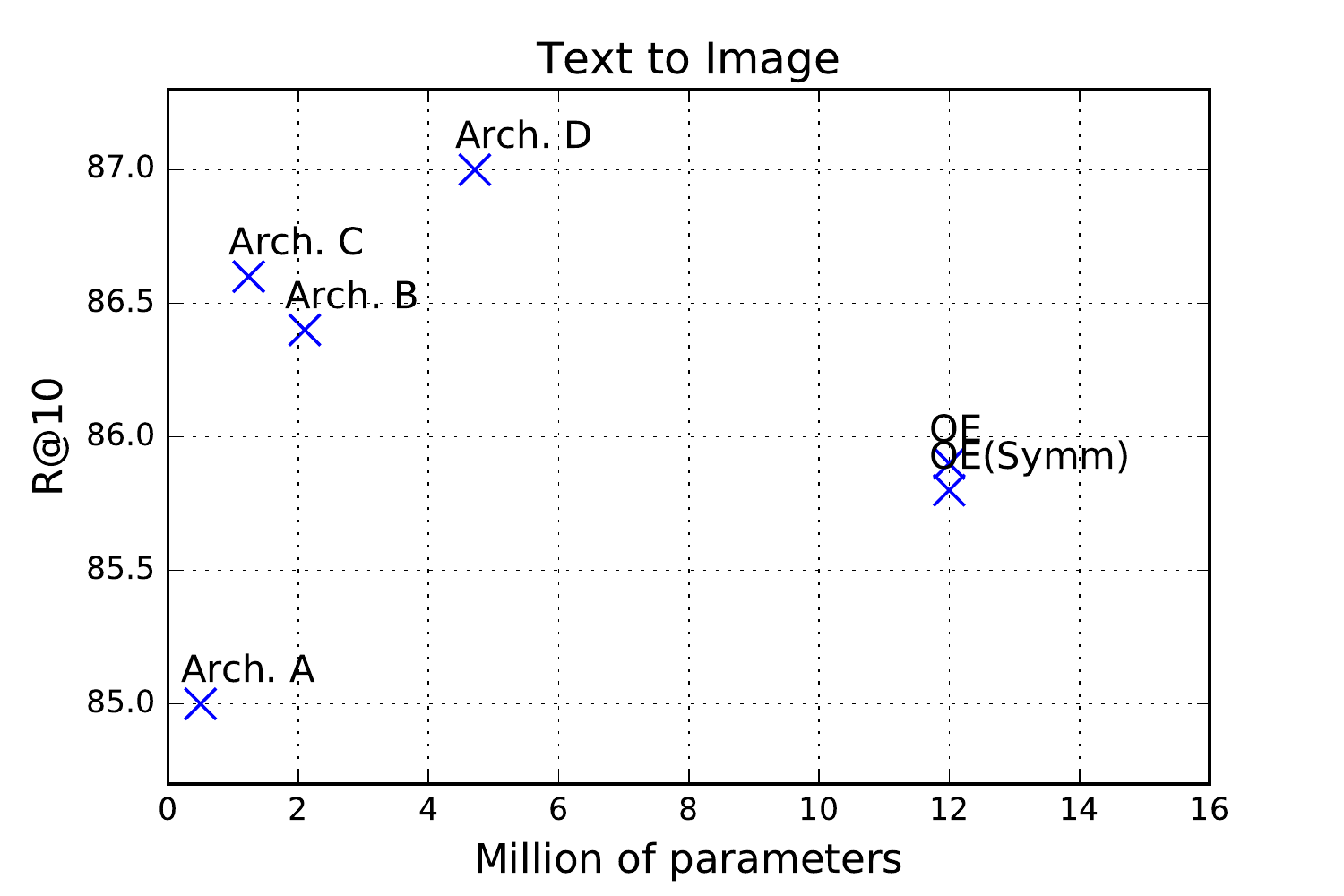}       
       \caption{Trade-off between model complexity (\#parameters) and predictive performance ($R@10$) in the test set. Results are for both image-to-text (above) and text-to-image retrieval (below). }
       \label{fig:parameter_comparison}
\end{figure}

Figure~\ref{fig:parameter_comparison} shows the trade-off analysis between model complexity and predictive performance of our models and the work in \citep{vendrov2015order}. The ideal position is in the upper-left position (largest recall and smallest number of parameters). For generating this visualization, we computed the number of trainable parameters for each method. Note that the word-embeddings are considered trainable parameters as well. For image-to-text retrieval both Arch.~C and~D are virtually in the same horizontal line than the architectures in \citep{vendrov2015order}, while requiring only a fraction of the processing needed by the baseline. Regarding the text-to-image results, note that Archs.-[B,C,D] present superior performance despite being up to ten-fold lighter (Arch.~C with 1.2M versus OE with 12M). 

\subsection{Qualitative Analysis}

\begin{figure*}[!htpb]
       \centering
       \includegraphics[scale=0.3]{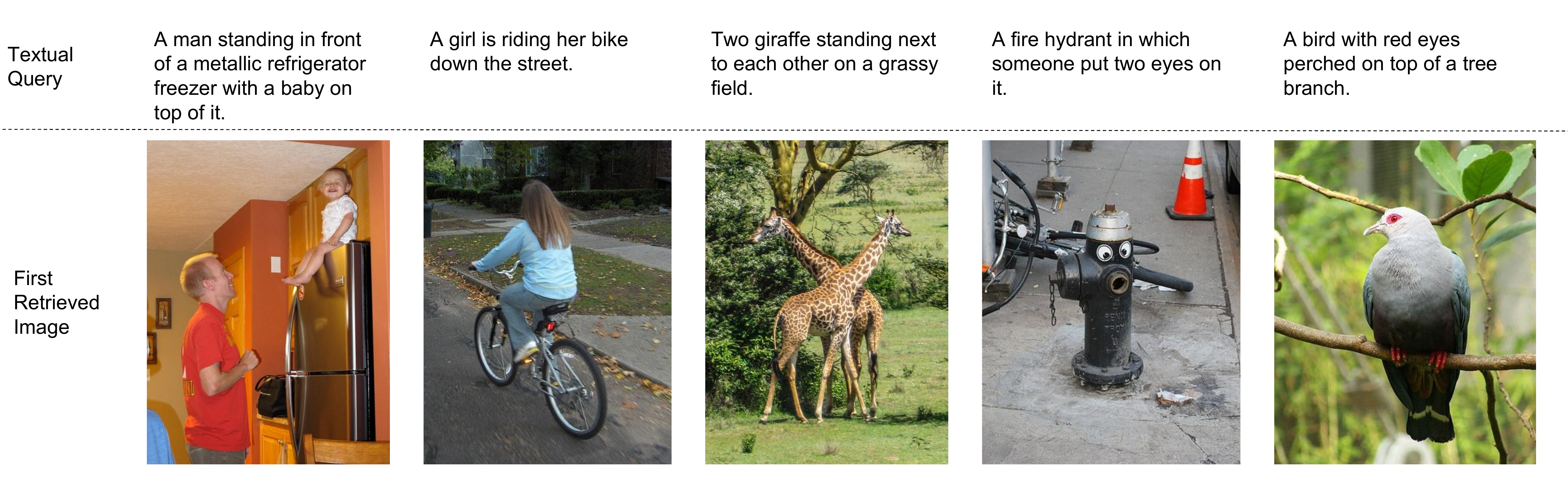}
       \caption{Text-to-image retrieval results. }
       \label{fig:t2i_results}
\end{figure*}

In this section we discuss qualitative results by analyzing images and text that were retrieved by using our best model, namely Arch.~D. Figure~\ref{fig:t2i_results} shows cases of image retrieval from textual queries. This is equivalent to image search taking into consideration the semantics of the text. Note that by using our strategy, it is possible to recover images from highly-detailed queries. For instance, our method is capable of learning spatial relations, objects, differentiate human genre (\textit{``a man is standing..."}, \textit{``a girl is riding..."}), age-related information (\textit{``baby is on top of..."}), counts (\textit{``two giraffes..."}), and also complex relationships (\textit{``a fire hydrant with two eyes..."}, or \textit{``a bird with red eyes on top of a tree..."}).

\begin{figure*}[!htpb]
       \centering
       \includegraphics[scale=0.3]{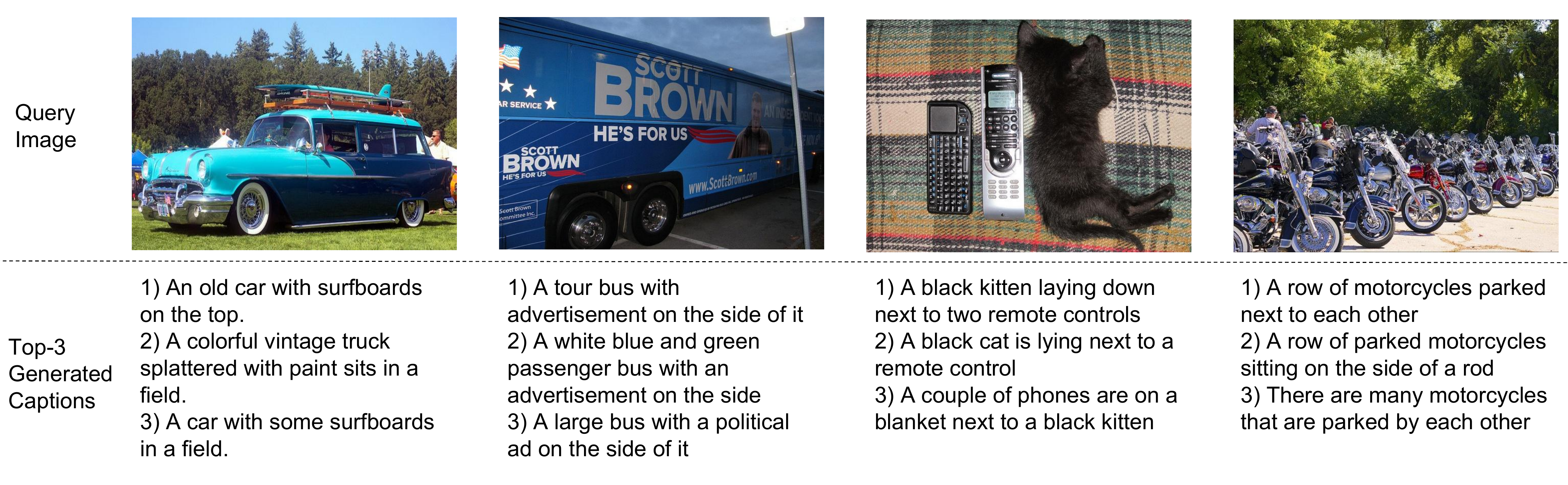}
       \caption{Image-to-text retrieval results. }
       \label{fig:i2t_results}
\end{figure*}

Figure~\ref{fig:i2t_results} depicts qualitative results for image-to-text retrieval. In this experiment, we performed queries based on images to retrieve the most similar captions in the multimodal embedding space. We show the three most-correlated captions indicated by our model for four randomly-chosen images. Note that, in all images, at least two ground-truth captions were retrieved. We did not highlight the wrongly-retrieved texts on purpose so we give the reader a chance to find them.

\begin{figure*}[!htpb]
       \centering
       \includegraphics[scale=0.33]{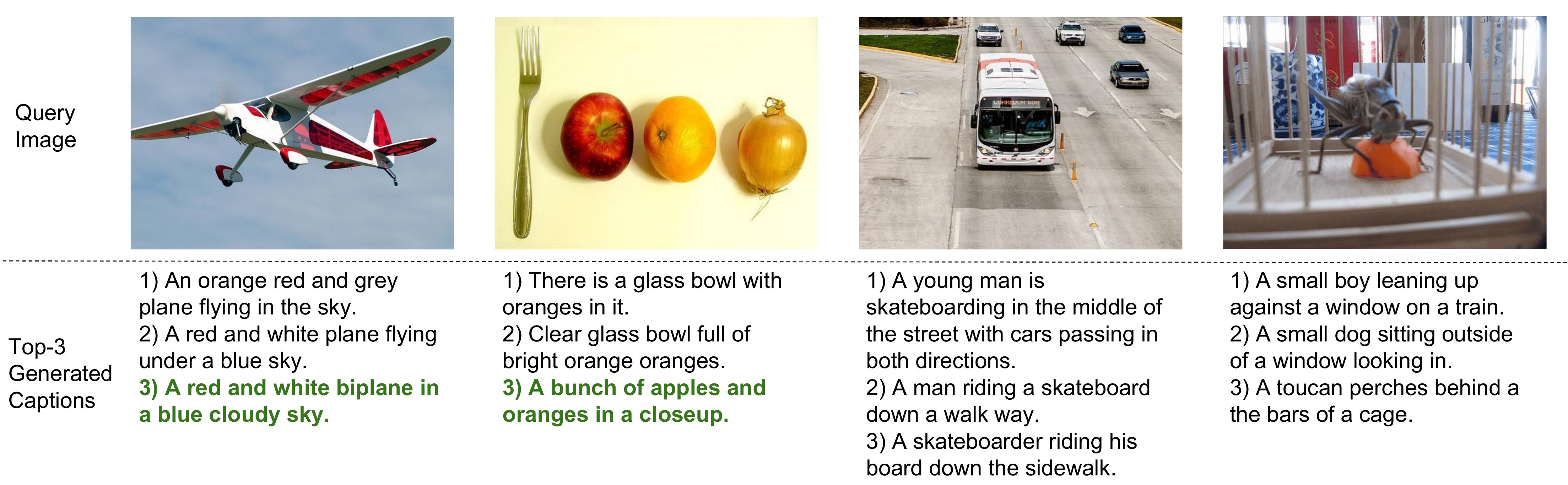}
       \caption{Failure cases of image-to-text retrieval. }
       \label{fig:it2_failures}
\end{figure*}

Finally, in Figure~\ref{fig:it2_failures} we show failure cases of image-to-text retrieval. We display the ground-truth captions in green (when recovered). Captions retrieved in the first image are reasonable, given that the model only confused the airplane colors. Descriptions for the second image detail the presence of oranges and a nonexistent glass bowl. Finally, the third and fourth image descriptions are interesting cases where the retrieved texts are plain wrong, probably due to the large variety of images/descriptions in the dataset. Particularly, the fourth image (insect inside a cage) depicts a very unusual scene.

\section{Related Work}\label{sec:rw}

Recent studies have proposed methods for bidirectional image and sentence retrieval \citep{kiros2014visual, kiros2014multimodal,karpathy2015deep,fv2015, SkipThoughtVectors,Socher2014, vendrov2015order} and automatic image captioning \citep{donahue2015long,xu2015show,show_and_tell,chen2015mind}.

\citet{kiros2014visual,kiros2014multimodal} propose a novel approach to deal with the problem of image caption generation by using an encoder-decoder architecture. For the encoding part of the architecture, a jointly-trained image-sentence embedding is generated, where sentences are encoded using GRU networks. Their approach is designed to map images and sentences onto the same embedding space so that correlated image-caption vectors present a high cosine similarity score. In addition, they also propose a language model for generating captions based on structure and content. 

\citet{karpathy2015deep} propose an architecture that makes use of  features from detection/localization-based systems. The idea is aligning image regions with a proper sentence fragment. This architecture learns inter-modal correspondences between image and sentence fragments, despite discarding global relations.

\citet{fv2015} present new Fisher Vectors that were derived from Hybrid Gaussian-Laplacian Mixture Models in order to provide sentence representations. They use Fisher Vectors as a pooling strategy of word-embeddings for representing text. Such representations are used in Laplacian Mixture Models trained with an expectation-maximization algorithm. 

\citet{SkipThoughtVectors} introduce a sentence embedding strategy, namely \textit{skip-thought vectors}, which is trained with no supervision over a very large book corpus to learn generic sentence representations. They build an encoder-decoder architecture that aims to reconstruct the surrounding sentences of a given encoded passage in a book. The authors encourage the use of a combined architecture of three skip-thought based models trained with different strategies (combine-skip-thought). This ensemble approach often leads to better results, though largely increasing the computational requirements. 

\citet{mcnn2015} propose a multimodal convolutional neural network for learning a matching score between image representations and text represented by sequences of word-embeddings, by jointly convolving the word-embeddings and image features. The learned similarity score predicts whether a pair is correlated or not. Architectural variations are explored and the best results are achieved with an ensemble version. 

Finally, \citet{vendrov2015order} propose the sentence order-embeddings. Those embeddings aim to preserve the partial order structure of a visual-semantic hierarchy. It allows learning ordered representation by applying order-penalties, and they show that asymmetric measures are better suited for image-sentence retrieval tasks. Their architecture is virtually the same of the one introduced in \citep{kiros2014visual}, with only the loss function being changed to consider the order violations. 

\section{Conclusions}\label{sec:conclusions}
We presented a simpler and faster architecture capable of learning textual embedding based on raw characters and order-embeddings for image-text alignment. Even though it is conceptually a much simpler architecture than those found in related work, our approach is capable of achieving state-of-the-art results in both text-to-image and image-to-text tasks. A promising future work direction is to analyze the performance of our approach in problems such as multilingual data analysis and caption generation for videos.

\section*{Acknowledgments}

We would like to thank Motorola Mobility, Google, and Brazilian research agencies CNPq, CAPES, and FAPERGS for funding this research.

\bibliographystyle{model2-names}
\bibliography{reference}

\begin{thebibliography}{22}
\expandafter\ifx\csname natexlab\endcsname\relax\def\natexlab#1{#1}\fi
\providecommand{\url}[1]{\texttt{#1}}
\providecommand{\href}[2]{#2}
\providecommand{\path}[1]{#1}
\providecommand{\DOIprefix}{doi:}
\providecommand{\ArXivprefix}{arXiv:}
\providecommand{\URLprefix}{URL: }
\providecommand{\Pubmedprefix}{pmid:}
\providecommand{\doi}[1]{\href{http://dx.doi.org/#1}{\path{#1}}}
\providecommand{\Pubmed}[1]{\href{pmid:#1}{\path{#1}}}
\providecommand{\bibinfo}[2]{#2}
\ifx\xfnm\relax \def\xfnm[#1]{\unskip,\space#1}\fi
\bibitem[{Chen and Lawrence~Zitnick(2015)}]{chen2015mind}
\bibinfo{author}{Chen, X.}, \bibinfo{author}{Lawrence~Zitnick, C.},
  \bibinfo{year}{2015}.
\newblock \bibinfo{title}{Mind's eye: A recurrent visual representation for
  image caption generation}, in: \bibinfo{booktitle}{Proceedings of the IEEE
  Conference on Computer Vision and Pattern Recognition (CVPR 2015)}, pp.
  \bibinfo{pages}{2422--2431}.
\bibitem[{Chung et~al.(2015)Chung, Gulcehre, Cho and Bengio}]{Chung2015}
\bibinfo{author}{Chung, J.}, \bibinfo{author}{Gulcehre, C.},
  \bibinfo{author}{Cho, K.}, \bibinfo{author}{Bengio, Y.},
  \bibinfo{year}{2015}.
\newblock \bibinfo{title}{Gated feedback recurrent neural networks}, in:
  \bibinfo{booktitle}{International Conference on Machine Learning (ICML
  2015)}, pp. \bibinfo{pages}{2067--2075}.
\bibitem[{Donahue et~al.(2015)Donahue, Anne~Hendricks, Guadarrama, Rohrbach,
  Venugopalan, Saenko and Darrell}]{donahue2015long}
\bibinfo{author}{Donahue, J.}, \bibinfo{author}{Anne~Hendricks, L.},
  \bibinfo{author}{Guadarrama, S.}, \bibinfo{author}{Rohrbach, M.},
  \bibinfo{author}{Venugopalan, S.}, \bibinfo{author}{Saenko, K.},
  \bibinfo{author}{Darrell, T.}, \bibinfo{year}{2015}.
\newblock \bibinfo{title}{Long-term recurrent convolutional networks for visual
  recognition and description}, in: \bibinfo{booktitle}{Proceedings of the IEEE
  conference on computer vision and pattern recognition}, pp.
  \bibinfo{pages}{2625--2634}.
\bibitem[{Hochreiter and Schmidhuber(1997)}]{Hochreiter1997}
\bibinfo{author}{Hochreiter, S.}, \bibinfo{author}{Schmidhuber, J.},
  \bibinfo{year}{1997}.
\newblock \bibinfo{title}{Long short-term memory}.
\newblock \bibinfo{journal}{Neural Comput.} \bibinfo{volume}{9},
  \bibinfo{pages}{1735--1780}.
\bibitem[{Karpathy and Fei-Fei(2015)}]{karpathy2015deep}
\bibinfo{author}{Karpathy, A.}, \bibinfo{author}{Fei-Fei, L.},
  \bibinfo{year}{2015}.
\newblock \bibinfo{title}{Deep visual-semantic alignments for generating image
  descriptions}, in: \bibinfo{booktitle}{Proceedings of the IEEE Conference on
  Computer Vision and Pattern Recognition (CVPR 2015)}, pp.
  \bibinfo{pages}{3128--3137}.
\bibitem[{Kingma and Ba(2014)}]{Kingma2014}
\bibinfo{author}{Kingma, D.P.}, \bibinfo{author}{Ba, J.}, \bibinfo{year}{2014}.
\newblock \bibinfo{title}{Adam: {A} method for stochastic optimization}, in:
  \bibinfo{booktitle}{International Conference on Learning Representations}.
\bibitem[{Kiros et~al.(2014a)Kiros, Salakhutdinov and
  Zemel}]{kiros2014multimodal}
\bibinfo{author}{Kiros, R.}, \bibinfo{author}{Salakhutdinov, R.},
  \bibinfo{author}{Zemel, R.}, \bibinfo{year}{2014}a.
\newblock \bibinfo{title}{Multimodal neural language models}, in:
  \bibinfo{booktitle}{Proceedings of the 31st International Conference on
  Machine Learning (ICML-14)}, pp. \bibinfo{pages}{595--603}.
\bibitem[{Kiros et~al.(2014b)Kiros, Salakhutdinov and Zemel}]{kiros2014visual}
\bibinfo{author}{Kiros, R.}, \bibinfo{author}{Salakhutdinov, R.},
  \bibinfo{author}{Zemel, R.S.}, \bibinfo{year}{2014}b.
\newblock \bibinfo{title}{Unifying visual-semantic embeddings with multimodal
  neural language models}.
\newblock \bibinfo{journal}{CoRR} \bibinfo{volume}{abs/1411.2539}.
\newblock \URLprefix \url{http://arxiv.org/abs/1411.2539}.
\bibitem[{Kiros et~al.(2015)Kiros, Zhu, Salakhutdinov, Zemel, Urtasun, Torralba
  and Fidler}]{SkipThoughtVectors}
\bibinfo{author}{Kiros, R.}, \bibinfo{author}{Zhu, Y.},
  \bibinfo{author}{Salakhutdinov, R.R.}, \bibinfo{author}{Zemel, R.},
  \bibinfo{author}{Urtasun, R.}, \bibinfo{author}{Torralba, A.},
  \bibinfo{author}{Fidler, S.}, \bibinfo{year}{2015}.
\newblock \bibinfo{title}{Skip-thought vectors}, in:
  \bibinfo{booktitle}{Advances in Neural Information Processing Systems (NIPS
  2015)}, pp. \bibinfo{pages}{3294--3302}.
\bibitem[{Klein et~al.(2015)Klein, Lev, Sadeh and Wolf}]{fv2015}
\bibinfo{author}{Klein, B.}, \bibinfo{author}{Lev, G.}, \bibinfo{author}{Sadeh,
  G.}, \bibinfo{author}{Wolf, L.}, \bibinfo{year}{2015}.
\newblock \bibinfo{title}{Associating neural word embeddings with deep image
  representations using fisher vectors}, in: \bibinfo{booktitle}{2015 IEEE
  Conference on Computer Vision and Pattern Recognition (CVPR)}, pp.
  \bibinfo{pages}{4437--4446}.
\newblock \DOIprefix\doi{10.1109/CVPR.2015.7299073}.
\bibitem[{Lin et~al.(2014)Lin, Maire, Belongie, Bourdev, Girshick, Hays,
  Perona, Ramanan, Doll{\'{a}}r and Zitnick}]{coco}
\bibinfo{author}{Lin, T.}, \bibinfo{author}{Maire, M.},
  \bibinfo{author}{Belongie, S.J.}, \bibinfo{author}{Bourdev, L.D.},
  \bibinfo{author}{Girshick, R.B.}, \bibinfo{author}{Hays, J.},
  \bibinfo{author}{Perona, P.}, \bibinfo{author}{Ramanan, D.},
  \bibinfo{author}{Doll{\'{a}}r, P.}, \bibinfo{author}{Zitnick, C.L.},
  \bibinfo{year}{2014}.
\newblock \bibinfo{title}{Microsoft {COCO:} common objects in context}, in:
  \bibinfo{booktitle}{European Conference on Computer Vision (ECCV 2014)}.
\bibitem[{Ma et~al.(2015)Ma, Lu, Shang and Li}]{mcnn2015}
\bibinfo{author}{Ma, L.}, \bibinfo{author}{Lu, Z.}, \bibinfo{author}{Shang,
  L.}, \bibinfo{author}{Li, H.}, \bibinfo{year}{2015}.
\newblock \bibinfo{title}{Multimodal convolutional neural networks for matching
  image and sentence}, in: \bibinfo{booktitle}{International Conference on
  Computer Vision (2015)}.
\bibitem[{Mao et~al.(2014)Mao, Xu, Yang, Wang and Yuille}]{m_rnn}
\bibinfo{author}{Mao, J.}, \bibinfo{author}{Xu, W.}, \bibinfo{author}{Yang,
  Y.}, \bibinfo{author}{Wang, J.}, \bibinfo{author}{Yuille, A.L.},
  \bibinfo{year}{2014}.
\newblock \bibinfo{title}{{Deep Captioning with Multimodal Recurrent Neural
  Networks (m-RNN)}}.
\newblock \bibinfo{journal}{CoRR} \bibinfo{volume}{abs/1412.6632}.
\newblock \URLprefix \url{http://arxiv.org/abs/1412.6632}.
\bibitem[{Mesnil et~al.(2013)Mesnil, He, Deng and
  Bengio}]{mesnil2013investigation}
\bibinfo{author}{Mesnil, G.}, \bibinfo{author}{He, X.}, \bibinfo{author}{Deng,
  L.}, \bibinfo{author}{Bengio, Y.}, \bibinfo{year}{2013}.
\newblock \bibinfo{title}{Investigation of recurrent-neural-network
  architectures and learning methods for spoken language understanding.}, in:
  \bibinfo{booktitle}{Interspeech}, pp. \bibinfo{pages}{3771--3775}.
\bibitem[{Mikolov et~al.(2013)Mikolov, Sutskever, Chen, Corrado and
  Dean}]{Mikolov2013}
\bibinfo{author}{Mikolov, T.}, \bibinfo{author}{Sutskever, I.},
  \bibinfo{author}{Chen, K.}, \bibinfo{author}{Corrado, G.},
  \bibinfo{author}{Dean, J.}, \bibinfo{year}{2013}.
\newblock \bibinfo{title}{Distributed representations of words and phrases and
  their compositionality}, in: \bibinfo{booktitle}{Advances on Neural
  Information Processing Systems (NIPS 2013)}, \bibinfo{address}{USA}. pp.
  \bibinfo{pages}{3111--3119}.
\bibitem[{Russakovsky et~al.(2015)Russakovsky, Deng, Su, Krause, Satheesh, Ma,
  Huang, Karpathy, Khosla, Bernstein, Berg and Fei-Fei}]{Russakovsky2015}
\bibinfo{author}{Russakovsky, O.}, \bibinfo{author}{Deng, J.},
  \bibinfo{author}{Su, H.}, \bibinfo{author}{Krause, J.},
  \bibinfo{author}{Satheesh, S.}, \bibinfo{author}{Ma, S.},
  \bibinfo{author}{Huang, Z.}, \bibinfo{author}{Karpathy, A.},
  \bibinfo{author}{Khosla, A.}, \bibinfo{author}{Bernstein, M.},
  \bibinfo{author}{Berg, A.C.}, \bibinfo{author}{Fei-Fei, L.},
  \bibinfo{year}{2015}.
\newblock \bibinfo{title}{{ImageNet Large Scale Visual Recognition Challenge}}.
\newblock \bibinfo{journal}{International Journal of Computer Vision}
  \bibinfo{volume}{115}, \bibinfo{pages}{211--252}.
\bibitem[{Simonyan and Zisserman(2015)}]{Simonyan2014}
\bibinfo{author}{Simonyan, K.}, \bibinfo{author}{Zisserman, A.},
  \bibinfo{year}{2015}.
\newblock \bibinfo{title}{Very deep convolutional networks for large-scale
  image recognition}, in: \bibinfo{booktitle}{International Conference on
  Learning Representations (ICLR 2015)}.
\bibitem[{Socher et~al.(2014)Socher, Karpathy, Le, Manning and Ng}]{Socher2014}
\bibinfo{author}{Socher, R.}, \bibinfo{author}{Karpathy, A.},
  \bibinfo{author}{Le, Q.V.}, \bibinfo{author}{Manning, C.D.},
  \bibinfo{author}{Ng, A.Y.}, \bibinfo{year}{2014}.
\newblock \bibinfo{title}{Grounded compositional semantics for finding and
  describing images with sentences}.
\newblock \bibinfo{journal}{{Transactions of the Association for Computational
  Linguistics (TACL)}} \bibinfo{volume}{2}, \bibinfo{pages}{207--218}.
\bibitem[{Vendrov et~al.(2016)Vendrov, Kiros, Fidler and
  Urtasun}]{vendrov2015order}
\bibinfo{author}{Vendrov, I.}, \bibinfo{author}{Kiros, R.},
  \bibinfo{author}{Fidler, S.}, \bibinfo{author}{Urtasun, R.},
  \bibinfo{year}{2016}.
\newblock \bibinfo{title}{Order-embeddings of images and language}, in:
  \bibinfo{booktitle}{International Conference on Learning Representations
  (ICLR 2016)}.
\bibitem[{Venugopalan et~al.(2016)Venugopalan, Hendricks, Rohrbach, Mooney,
  Darrell and Saenko}]{noc}
\bibinfo{author}{Venugopalan, S.}, \bibinfo{author}{Hendricks, L.A.},
  \bibinfo{author}{Rohrbach, M.}, \bibinfo{author}{Mooney, R.J.},
  \bibinfo{author}{Darrell, T.}, \bibinfo{author}{Saenko, K.},
  \bibinfo{year}{2016}.
\newblock \bibinfo{title}{Captioning images with diverse objects}.
\newblock \bibinfo{journal}{CoRR} \bibinfo{volume}{abs/1606.07770}.
\newblock \URLprefix \url{http://arxiv.org/abs/1606.07770}.
\bibitem[{Vinyals et~al.(2015)Vinyals, Toshev, Bengio and
  Erhan}]{show_and_tell}
\bibinfo{author}{Vinyals, O.}, \bibinfo{author}{Toshev, A.},
  \bibinfo{author}{Bengio, S.}, \bibinfo{author}{Erhan, D.},
  \bibinfo{year}{2015}.
\newblock \bibinfo{title}{Show and tell: A neural image caption generator}, in:
  \bibinfo{booktitle}{IEEE Internatinoal Conference on Computer Vision and
  Pattern Recognition (CVPR 2015)}, pp. \bibinfo{pages}{3156--3164}.
\bibitem[{Xu et~al.(2015)Xu, Ba, Kiros, Cho, Courville, Salakhudinov, Zemel and
  Bengio}]{xu2015show}
\bibinfo{author}{Xu, K.}, \bibinfo{author}{Ba, J.}, \bibinfo{author}{Kiros,
  R.}, \bibinfo{author}{Cho, K.}, \bibinfo{author}{Courville, A.},
  \bibinfo{author}{Salakhudinov, R.}, \bibinfo{author}{Zemel, R.},
  \bibinfo{author}{Bengio, Y.}, \bibinfo{year}{2015}.
\newblock \bibinfo{title}{Show, attend and tell: Neural image caption
  generation with visual attention}, in: \bibinfo{booktitle}{International
  Conference on Machine Learning (ICML 2015)}, pp. \bibinfo{pages}{2048--2057}.

\end{thebibliography}

\end{document}